\documentclass[11pt]{article}
\usepackage{acl2017}
\usepackage{latexsym}
\usepackage{graphicx}
\usepackage{}
\usepackage{amsfonts}
\usepackage{url}
\usepackage{mwe}
\usepackage{times}
\usepackage{url}
\usepackage{latexsym}
\usepackage{multirow}
\usepackage{microtype}
\usepackage{enumitem}
\usepackage{amsmath}
\usepackage{lipsum}
\usepackage{epstopdf}
\usepackage{xcolor}
\usepackage{tikz}
\usetikzlibrary{calc}
\usepackage{ragged2e}
\usepackage{subfigure}
\usepackage[compact]{titlesec}
\titlespacing{\section}{0pt}{2ex}{1ex}
\titlespacing{\subsection}{0pt}{1ex}{0ex}
\titlespacing{\subsubsection}{0pt}{0.5ex}{0ex}
\usepackage[export]{adjustbox}[2011/08/13]
\aclfinalcopy % Uncomment this line for the final submission

\title{Diversity driven Attention Model for Query-based Abstractive Summarization}

\author{Preksha Nema$^\dagger$ \hspace{0.1cm} Mitesh M. Khapra$^\dagger$ \hspace{0.1cm} Anirban Laha$^{*\dagger}$ \hspace{0.1cm} Balaraman Ravindran$^\dagger$\\
  $^\dagger$Indian Institute of Technology Madras, India \\ 
  $^*$ IBM Research India\\
  {\tt \{preksha,miteshk\}@cse.iitm.ac.in} \\ {\tt anirlaha@in.ibm.com} \hspace{0.1cm}  {\tt ravi@cse.iitm.ac.in} \\}

\date{}

\begin{document}\fontsize{11.5}{14}\rm
\maketitle
\begin{abstract}

Abstractive summarization aims to generate a shorter version of the document covering all the salient points in a compact and coherent fashion. On the other hand, query-based summarization highlights those points that are relevant in the context of a given query. The encode-attend-decode paradigm has achieved notable success in machine translation, extractive summarization, dialog systems, etc. But it suffers from the drawback of generation of repeated phrases. In this work we propose a model for the query-based summarization task based on the encode-attend-decode paradigm with two key additions (i) a query attention model (in addition to document attention model) which learns to focus on different portions of the query at different time steps (instead of using a static representation for the query) and (ii) a new diversity based attention model which aims to alleviate the problem of repeating phrases in the summary. In order to enable the testing of this model we introduce a new query-based summarization dataset building on debatepedia. Our experiments show that with these two additions the proposed model clearly outperforms vanilla encode-attend-decode models with a gain of 28\% (absolute) in ROUGE-L scores.
\end{abstract}

%establish initial baselines on this dataset by suitably extending existing abstractive summarization models to incorporate a given query.

\section{Introduction}
Over the past few years neural models based on the encode-attend-decode \cite{bahdanau2014neural} paradigm have shown great success in various natural language generation (NLG) tasks such as machine translation \cite{bahdanau2014neural}, abstractive summarization (\cite{rush2015neural},\cite{nallapati2016abstractive}) dialog \cite{li2016persona}, \textit{etc}. One such NLG problem which has not received enough attention in the past is query based abstractive text summarization where the aim is to generate the summary of a document in the context of a query. In general, abstractive summarization, aims to cover all the salient points of a document in a compact and coherent fashion. On the other hand, query focused summarization highlights those points that are relevant in the context of the query. Thus given a document on ``the super bowl'',  the query ``How was the half-time show?'', would result in a summary that would not cover the actual game itself.

Note that there has been some work on query based extractive summarization in the past where the aim is to simply extract the most salient sentence(s) from a document and treat these as a summary. There is no natural language generation involved. Since, we were interested in abstractive (as opposed to extractive) summarization we created a new dataset based on debatepedia. This dataset contains triplets of the form (query, document, summary). Further, each summary is abstractive and not extractive in the sense that the summary does not necessarily comprise of a sentence which is simply copied from the original document. 

Using  this dataset as a testbed, we focus on a recurring problem in models based on the encode-attend-decode paradigm. Specifically, it is observed that the summaries produced by such models contain repeated phrases. Table \ref{rp} shows a few such examples of summaries generated by such a model when trained on this new dataset. This problem has also been reported by \cite{chen2016distraction} in the context of summarization and by \cite{sankaran2016temporal} in the context of machine translation.

\begin{table}
\begin{tabular}{|p{0.45\textwidth}|}
\hline 
%\textbf{Repeated Phrases Problem}\\
\hline
\textbf{Document Snippet:}
 The ``natural death'' alternative to euthanasia is not keeping someone alive via life support until they die on life support. That would, indeed, be unnatural. The natural alternative is, instead, to allow them to die off of life support.  \\
\textbf{Query}: Is euthanasia better than withdrawing life support (non-treatment)? \\
\textbf{Ground Truth Summary}:
 The alternative to euthanasia is a natural death without life support.   \\

\textbf{Predicted Summary}:
the large to euthanasia is a natural death \textbf{life life} use \\
\hline
\textbf{Document Snippet:}
Legalizing same-sex marriage would also be a recognition of basic American principles, and would represent the culmination of our nation's commitment to equal rights. It is, some have said, the last major civil-rights milestone yet to be surpassed in our two-century struggle to attain the goals we set for this nation at its formation.  \\
\textbf{Query:} Is gay marriage a civil right?\\
\textbf{Ground Truth Summary}:
 Gay marriage is a fundamental equal right. \\
\textbf{Predicted Summary}:
gay marriage is a appropriate \textbf{right right} \\
\hline

\end{tabular}
\\
\caption{Examples showing repeated words in the output of encoder-decoder models}
\label{rp}
\end{table}

We first provide an intuitive explanation for this problem and then propose a solution for alleviating it. A typical encode-attend-decode model first computes a vectorial representation for the document and the query and then produces a contextual summary one word at a time. Each word is produced by feeding a new context vector to the decoder at each time step by attending to different parts of the document and query. If the decoder produces the same word or phrase repeatedly then it could mean that the context vectors fed to the decoder at these time steps are very similar.

We propose a model which explicitly prevents this by ensuring that successive context vectors are orthogonal to each other. Specifically, we subtract out any component that the current context vector has in the direction of the previous context vector. Notice that, we do not require the current context vector to be orthogonal to all previous context vectors but just its immediate predecessor. This enables the model to attend to words repeatedly if required later in the process. To account for the complete history (or all previous context vectors) we also propose an extension of this idea where we pass the sequence of context vectors through a LSTM \cite{hochreiter1997long} and ensure that the current state produced by the LSTM is orthogonal to the history. At each time step, the state of the LSTM is then fed to the decoder to produce one word in the summary. 

Our contributions can be summarized as follows: (i) We propose a new dataset for query based abstractive summarization and evaluate encode-attend-decode models on this dataset (ii) We study the problem of repeating phrases in NLG in the context of this dataset and propose two solutions for countering this problem. We show that our method outperforms a vanilla encoder-decoder model with a gain of 28\% (absolute) in ROUGE-L score (iii) We also demonstrate that our method clearly outperforms a recent state of the art method proposed for handling the problem of repeating phrases with a gain of 7\% (absolute) in ROUGE-L scores (iv) We do a qualitative analysis of the results and show that our model indeed produces outputs with fewer repetitions.

\section{Related Work}
Summarization has been studied in the context of text (\cite{mani2001automatic}, \cite{das2007survey}, \cite{nenkova2012survey}) as well as speech (\cite{zhu2006comparing}, \cite{zhu2009summarizing}). A vast majority of this work has focused on extractive summarization where the idea is to construct a summary by selecting the most relevant sentences from the document (\cite{neto2002automatic}, \cite{erkan2004lexrank}, \cite{filippova2013overcoming}, \cite{colmenares2015heads}, \cite{riedhammer2010long}, \cite{ribeiro2013self}). There has been some work on abstractive summarization in the context of DUC-2003 and DUC-2004 contests \cite{zajic2004bbn}. We refer the reader to \cite{das2007survey} and \cite{nenkova2012survey} for an excellent survey of the field.

Recent research in abstractive summarization has focused on data driven neural models based on the encode-attend-decode paradigm \cite{bahdanau2014neural}.  For example, \cite{rush2015neural}, report state of the art results on the GigaWord and DUC corpus using such a model. Similarly, the work of \newcite{lopyrev2015generating} uses neural networks to generate news headline from short news stories. \newcite{chopra2016abstractive} extend the work of \newcite{rush2015neural} and report further improvements on the two datasets. \newcite{hu2015lcsts} introduced a dataset for Chinese short text summarization and evaluated a similar RNN encoder-decoder model on it.

One recurring problem in encoder-decoder models for NLG is that they  often repeat the same phrase/word multiple times in the summary (at the cost of both coherency and fluency). \newcite{sankaran2016temporal} study this problem in the context of MT and propose a temporal attention model which enforces the attention weights for successive time steps to be different from each other. Similarly, and more relevant to this work, \newcite{chen2016distraction} propose a distraction based attention model which maintains a history of attention vectors and context vectors. It then subtracts this history from the current attention and context vector. When evaluated on our dataset their method performs poorly. This could be because their method is very aggressive in dealing with the history (as explained later in the Experiments section). On the other hand, our method has a better way of handling history (by passing context vectors through an LSTM recurrent network) which gives us the flexibility to forget/retain some portions of the history and at the same time produce diverse context vectors at successive time steps.    

We evaluate our method in the context of query based abstractive summarization - a problem which has received almost no attention in the past due to unavailability of datasets. We create a new dataset for this task and show that our method indeed produces better output by reducing the number of repeated phrases produced by encoder decoder models. 

\begin{figure*}[!tbh]
\begin{minipage}{0.5\textwidth}
\includegraphics[width=\textwidth]{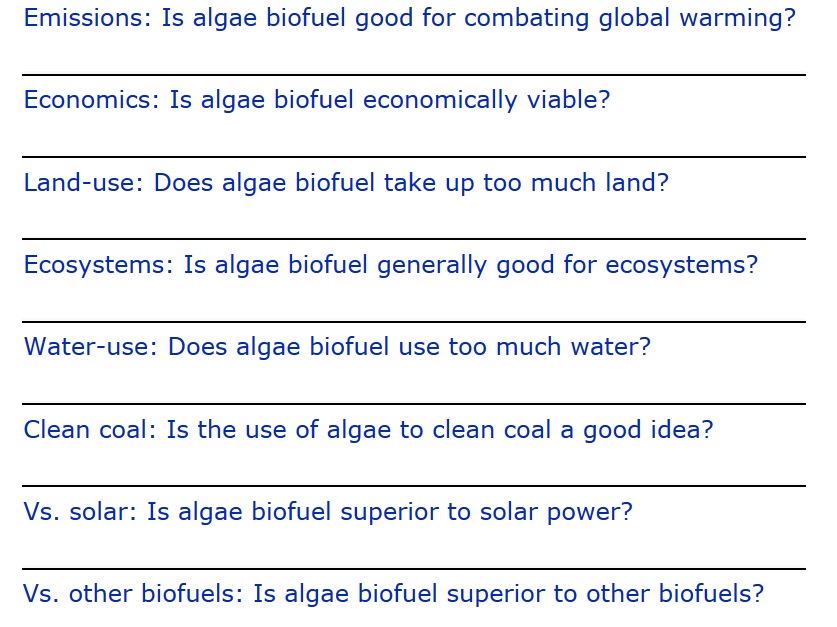}
\caption{Queries associated with the topic ``algae biofuel''}
\label{q}
\end{minipage}
\begin{minipage}{0.5\textwidth}
\includegraphics[width=\textwidth]{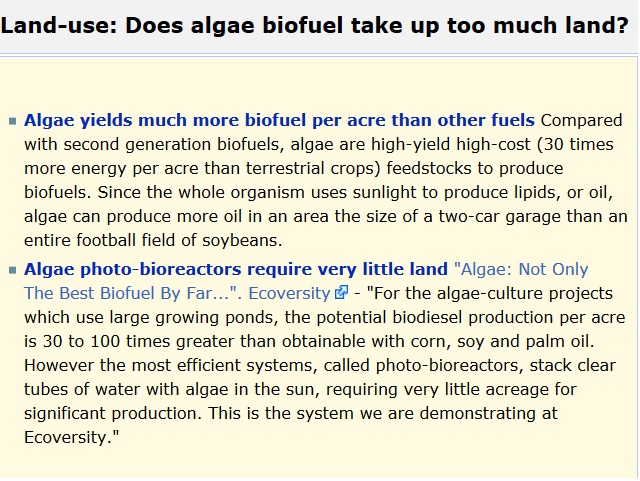}
\caption{Documents and summaries for a given query}
\label{d}
\end{minipage}

\end{figure*}

\section{Dataset}
As mentioned earlier, there are no existing datasets for query based abstractive summarization. We create such a dataset from Debatepedia an encyclopedia of pro and con arguments and quotes on critical debate topics. There are 663 debates in the corpus (we have considered only those debates which have at least one query with one document). These 663 debates belong to 53 overlapping categories such as Politics, Law, Crime, Environment, Health, Morality, Religion, etc. A given topic can belong to more than one category. For example, the topic ``Eye for an Eye philosophy'' belongs to both ``Law'' as well as ``Morality''. %%CameraReady
The average number of queries per debate is 5 and the average number of documents per query is 4. Please refer to the dataset url\footnote{\url{https://github.com/PrekshaNema25/DiverstiyBasedAttentionMechanism}} for more details about number of debates per category.
%%Each of the debate topic has a set of queries associated with it.
%%For example, Figure \ref{q} shows the queries associated with the topic ``Algae Biofuel''. Each query has a set of documents and an abstractive summary associate with each document (see Figure \ref{d}). As is obvious from the example, the summary is an abstractive summary and not extracted directly from the document. We crawled 12695 such \{query, document, summary\} triples from debatepedia (these were all the triples that were available). Table \ref{table_ds} reports the average length of the query, summary and documents in this dataset. 

For example, Figure \ref{q} shows the queries associated with the topic ``Algae Biofuel''. It also lists the set of documents and an abstractive summary associated with each query. As is obvious from the example, the summary is an abstractive summary and not extracted directly from the document. We crawled 12695 such \{query, document, summary\} triples from debatepedia (these were all the triples that were available). Table \ref{table_ds} reports the average length of the query, summary and documents in this dataset. 

We used 10 fold cross validation for all our experiments. Each fold uses 80\% of the documents for training, 10\% for validation and 10\% for testing.

\begin{center}
\begin{table}
\begin{center}
\begin{tabular}{ccc}
\hline
\multicolumn{3}{c}{\textbf{Average number of words per}}\\
\hline
Document & Summary & Query\\
\hline
66.4 & 11.16 & 9.97\\
\hline
&\\
\end{tabular}
\end{center}
\caption{Average length of documents/queries/summaries in the dataset}
\label{table_ds}
\end{table}
\end{center}
\if 0
\begin{figure*}
%\includegraphics[width=\textwidth, height=9cm]{dyn.jpg}
\include{query}
\caption{Proposed encode-attend-decode model for Query based Abstractive Summarization with (i) query encoder (ii) document encoder (iii) }
\label{dyn}
\end{figure*}
\fi

\section{Proposed model}
Given a query $\mathbf{q} = q_1, q_2, ..., q_k$ containing $k$ words, a document $\mathbf{d} = d_1, d_2, ..., d_n$ containing $n$ words, the task is to generate a contextual summary $\mathbf{y} = y_1, y_2, ..., y_m$ containing $m$ words. This can be modeled as the problem of finding  a $\mathbf{y}^*$ that maximizes the probability $p(\mathbf{y}|\mathbf{q}, \mathbf{d})$ which can be further decomposed as:
\begin{align}
\label{eq0}
y^*  &= \arg\max_{y} \prod_{t=1}^{m} p(y_t| y_1, ..., y_{t-1}, \mathbf{q}, \mathbf{d})
\end{align}

We now describe a way of modeling  $p(y_t| y_1, ..., y_{t-1}, \mathbf{q}, \mathbf{d})$ using the neural encoder-attention-decoder paradigm. The proposed model contains the following components: (i) an encoder RNN for the query (ii) an encoder RNN for the document (iii) attention mechanism for the query (iv) attention mechanism for the document and (v) a decoder RNN. All the RNNs use a GRU cell.

\noindent \textbf{Encoder for the query:} We use a recurrent neural network with Gated Recurrent Units (GRU) for encoding the query. It reads the query $\mathbf{q} = q_1, q_2, ..., q_k$ from left to right and computes a hidden representation for each time-step as:
\begin{align}
h_i^q = \text{GRU}_q(h_{i-1}^q, e(q_i))
\end{align}
where $e(q_i) \in \mathbb{R}^d$ is the $d$-dimensional embedding of the query word $q_i$. 

\noindent \textbf{Encoder for the document:} This is similar to the query encoder and reads the document $\mathbf{d} = d_1, d_2, ..., d_n$ from left to right and computes a hidden representation for each time-step as: 
\begin{align}
h_i^d = \text{GRU}_d(h_{i-1}^d, e(d_i))
\end{align}
where $e(d_i) \in \mathbb{R}^d$ is the $d$-dimensional embedding of the document word $d_i$.

\noindent \textbf{Attention mechanism for the query :} At each time step, the decoder produces an output word by focusing on different portions of the query (document) with the help of a query (document) attention model. We first describe the query attention model which assigns weights $\alpha_{t,i}^q$ to each word in the query at each decoder timestep using the following equations.

\begin{align}
a_{t,i}^q &= v_q^T \tanh(W_qs_t + U_qh^q_i) \\
\alpha_{t,i}^q &= \frac{\text{exp}(a_{t,i}^q)}{\sum_{j=1}^{k} \text{exp}(a_{t,j}^q)}
\end{align}
where $s_t$ is the current state of the decoder at time step $t$ (we will see an exact formula for this soon). $W_q \in \mathbb{R}^{l_{2} \times l_{1}}$, $U_q \in \mathbb{R}^{l_{2} \times l_{2}}$, $v_q \in \mathbb{R}^{l_{2}}$, $l_{1}$ is the size of the decoder's hidden state, $l_{2}$ is both the size of $h_i^q$ and also the size of the final query representation at time step $t$, which is computed as:
\begin{align}
q_t &= \sum_{i=1}^{k} \alpha_{t,i}^q h_{i}^{q}
\end{align}

\noindent \textbf{Attention mechanism for the document :} We now describe the document attention model which assigns weights to each word in the document using the following equations.

\begin{align}
\label{eq2}
a_{t,i}^d &= v_d^T \tanh(W_ds_t + U_dh^d_i + Zq_t) \\
\nonumber \alpha_{t,i}^d &= \frac{\text{exp}(a_{t,i}^d)}{\sum_{j=1}^{n} \text{exp}(a_{t,j}^d)}
\end{align}
where $s_t$ is the current state of the decoder at time step $t$ (we will see an exact formula for this soon). $W_{d} \in \mathbb{R}^{l_{4} \times l_{1}}$, $U_d \in \mathbb{R}^{l_{4} \times l_{4}}$, $Z \in \mathbb{R}^{l_{4} \times l_{2}}$, $v_d \in \mathbb{R}^{l_{2}}$, $l_{4}$ is the size of $h_i^d$ and also the size of the final document representation $d_t$ which is passed to the decoder at time step $t$ as:
\begin{align}
\label{eq7}
d_t &= \sum_{i=1}^{n} \alpha_{t,i}^d h_{i}^{d}
\end{align}
Note that $d_t$ now encodes the relevant information from the document as well as the query (see Equation \eqref{eq2}) at time step $t$. We refer to this as the \textit{context vector} for the decoder.

\noindent \textbf{Decoder:} The hidden state of the decoder $s_t$ at each time $t$ is again computed using a GRU as follows:

\begin{align}
\label{eq4}
s_t = \text{GRU}_{dec}(s_{t-1}, [e(y_{t-1}), d_{t-1}])
\end{align}

\noindent where, $y_{t-1}$ gives a distribution over the vocabulary words at timestep $t-1$ and is computed as: 

\begin{align}
\label{eq5}
%y_t = \arg\max_{N}\{\text{softmax}(W_o f(W_{dec}s_t + V_{dec}d_t))\}
y_t = \text{softmax}(W_o f(W_{dec}s_t + V_{dec}d_t))
\end{align}

\noindent where $W_o \in \mathbb{R}^{N \times l_{1}}$, $W_{dec} \in \mathbb{R}^{l_{1} \times l_{1}}$, $V_{dec} \in \mathbb{R}^{l_{1} \times l_{4}}$, $N$ is the vocabulary size, $y_t$ is the final output of the model which defines a probability distribution over the output vocabulary. This is exactly the quantity defined in Equation \eqref{eq0} that we wanted to model ($p(y_t| y_1, ..., y_{t-1}, \mathbf{q}, \mathbf{d}$)). Further, note that, $e(y_{t-1})$ is the $d$-dimensional embedding of the word which has the highest probability under the distribution $y_{t-1}$. Also $[e(y_{t-1}), d_{t-1}]$ means a concatenation of the vectors $e(y_{t-1}), d_{t-1}$. We chose $f$ to be the identity function. 

\if 0
\begin{align}
\label{eq5}
y_t = \arg\max_{N}\{\text{softmax}(W_o f(W_{dec}s_t + V_{dec}d_t))\}
\end{align}
\fi 
 The model as  described above is an instantiation of the encoder-attention-decoder idea applied to query based abstractive summarization. As mentioned earlier (and demonstrated later through experiments), this model suffers from the problem of repeating the same phrase/word in the output. We now propose  a new attention model which we refer to as diversity based attention model to address this problem.

\begin{figure}[!tbh]
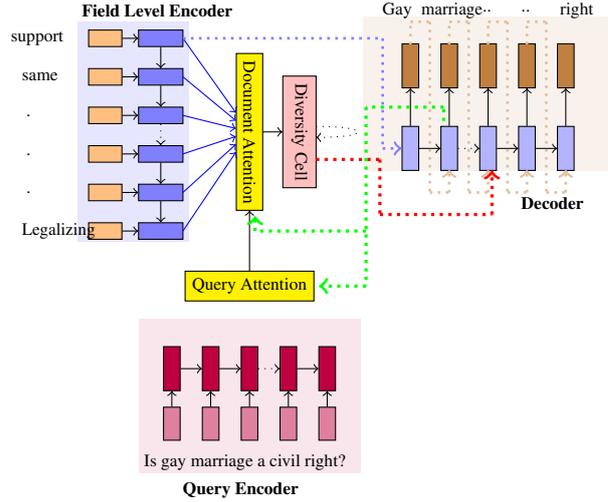


%\includegraphics[width=\textwidth, height=8cm]{dyndiv.jpg}
%\begin{minipage}{\textwidth}
\include{query}
\caption{Proposed model for Query based Abstractive Summarization with (i) query encoder (ii) document encoder (iii) query attention model (iv) diversity based document attention model and (v) decoder. The green and red arrows show the connections for timestep 3 of the decoder.}
\label{div}
%\end{minipage}
\end{figure}

\subsection{Diversity based attention model}
As hypothesized earlier, if the decoder produces the same phrase/word multiple times then it is possible that the context vectors being fed to the decoder at consecutive time steps are very similar. We propose four models ($\mathbf{D_1}$, $\mathbf{D_2}$, $\mathbf{SD_1}$, $\mathbf{SD_2}$) to directly address this problem.

\noindent\textbf{$\mathbf{D_1}$:} In this model, after computing $d_t$ as described in Equation \eqref{eq7}, we make it orthogonal to the context vector at time $t-1$:
\begin{align}
d_t^{'} = d_t - \frac{d_t^T d_{t-1}^{'}}{d_{t-1}^{'^T}d_{t-1}^{'}} d_{t-1}^{'}
\end{align}
\noindent\textbf{$\mathbf{SD_1}$:} The above model imposes a hard orthogonality constraint on the context vector($d_t^{'}$). We also propose a relaxed version of the above model which uses a gating parameter. This gating parameter decides what fraction of the previous context vector should be subtracted from the current context vector using the following equations:
\begin{align}
\nonumber \gamma_{t} &= W_{g}d_{t-1} + b_{g} \\
\nonumber d_t^{'} &= d_t - \gamma_{t} \frac{d_t^T d_{t-1}^{'}}{d_{t-1}^{'^T}d_{t-1}^{'}} d_{t-1}^{'}
\end{align}
where $W_{g} \in \mathbb{R}^{l_4 \times l_4}$, $b_{g} \in \mathbb{R}^{l_4}$, ${l_4}$ is the dimension of $d_t$ as defined in equation \eqref{eq7}.

\noindent\textbf{$\mathbf{D_2}$:} The above model only ensures that the current context vector is diverse w.r.t the previous context vector. It ignores all history before time step $t-1$. To account for the history, we treat successive context vectors as a sequence and use a modified LSTM cell   to compute the new state at each time step. Specifically, we use the following set of equations to compute a diverse context at time $t$:

\begin{align}
\label{eq12}
\nonumber i_t &= \sigma(W_{i}d_{t} + U_{i}h_{t-1} +b_{i})\\
\nonumber f_t &= \sigma(W_{f}d_{t} + U_{f}h_{t-1} +b_{f}) \\
\nonumber o_t &= \sigma(W_{o}d_{t} + U_{o}h_{t-1} +b_{o})\\
\nonumber \hat{c_t} &= \tanh(W_{c}d_{t} + U_{c}h_{t-1} +b_{c})\\
\nonumber {c_t} &= i_{t}\odot\hat{c_t} + f_{t}\odot c_{t-1} \\
c_t^{diverse} &= {c_t} - \frac{{c_t}^T c_{t-1}}{c_{t-1}^{T}c_{t-1}} c_{t-1}\\
\nonumber h_t &= o_{t}\odot \tanh(c_t^{diverse}) \\
\label{eq13} d_t^{'} &= h_t
\end{align}

where $W_{i}, W_{f}, W_{o}, W_{c} \in \mathbb{R}^{l_{5} \times l_{4}}$, $U_{i}, U_{f}, U_{o}, U_{c} \in \mathbb{R}^{l_{5} \times l_{4}}$, $d_{t}$ is the $l_{4}$-dimensional output of Equation (8); $l_{5}$ is number of hidden units in the LSTM cell. This final $d_t^{'}$ from Equation \eqref{eq13} is then used in Equation \eqref{eq4}. Note that Equation \eqref{eq12} ensures that state of the LSTM at time step $t$ is orthogonal to the previous history. Figure \ref{div} shows a pictorial representation of the model with a diversity LSTM cell.
%% CameraReady

\noindent\textbf{$\mathbf{SD_2}$:} This model again uses a relaxed version of the orthogonality constraint used in \textbf{$\mathbf{D_2}$}. Specifically, we define a gating parameter $g_t$ and replace \eqref{eq12} above by \eqref{eq14} as define below:
\begin{align}
\label{eq14}
\nonumber g_t &= \sigma(W_{g}d_{t} + U_{g}h_{t-1} + b_{o})\\
c_t^{diverse} &= {c_t} - {g_t}\frac{{c_t}^T c_{t-1}}{c_{t-1}^{T}c_{t-1}} c_{t-1}
\end{align}
where $W_{g} \in \mathbb{R}^{l_{5} \times l_{4}}$, $U_{g} \in \mathbb{R}^{l_{5} \times l_{4}}$

\section{Baseline Methods}
We compare with two recently proposed baseline diversity methods \cite{chen2016distraction} as described below. Note that these methods were proposed in the context of abstractive summarization (not query based abstractive summarization) and we adapt them for the task of query based abstractive summarization. Below we just highlight the key differences from our model in computing the context vector $d_t^{'}$ passed to the decoder.

\noindent\textbf{M1:} This model accumulates all the previous context vectors as $\sum_{j=1}^{t-1}d_j^{'}$ and incorporates this history while computing a diverse context vector:
\begin{align}
\label{eq15}
d_t^{'} = \tanh(W_cd_t - U_c \sum_{j=1}^{t-1}d_j^{'})
\end{align}
where $W_c, U_c \in \mathbb{R}^{l_{4} \times l_{4}}$ are diagonal matrices.
We then use this diversity driven context $d_t^{'}$ in Equation \eqref{eq4} and \eqref{eq5}.

\noindent\textbf{M2:} In this model, in addition to computing a diverse context as described in Equation \eqref{eq15}, the attention weights at each time step are also forced to be diverse from the attention weights at the previous time step. 
\begin{align}
\nonumber \alpha_{t,i}^{'} &= v_a^T \tanh(W_as_t^{'} + U_ad_t - b_a\sum_{j=1}^{t-1} \alpha_{j,i}^{'})
\end{align}
where $W_{a} \in \mathbb{R}^{l_{1} \times l_{1}}$, $U_{a} \in \mathbb{R}^{l_{1} \times  l_{4}}$, $b_{a}, v_{a} \in \mathbb{R}^{l_1}$, $l_{1}$ is the number of hidden units in the decoder GRU. Once again, they maintain a history of attention weights and compute a diverse attention vector by subtracting the history from the current attention vector. 
%In practice, we observed that both these models perform badly on our dataset. We give plausible explanations for this in the Discussions section.

\section{Experimental Setup}
We evaluate our models on the dataset described in section 3. Note that there are no prior baselines on query based abstractive summarization so we could only compare with different variations of the encoder decoder models as described above. Further, we compare our diversity based attention models with existing models for diversity by suitably adapting them to this problem as described earlier. Specifically, we compare the performance of the following models:

\begin{itemize}
\item \textbf{Vanilla e-a-d}: This is the vanilla encoder-attention-decoder model adapted to the problem of abstractive summarization. It contains the following components (i) document encoder (ii) document attention model (iii) decoder. It does not contain an encoder or attention model for the query. This helps us understand the importance of the query.
\vspace{-0.05in}
\item \textbf{$\mathbf{Query}_{enc}$}: This model contains the query encoder in addition to the three components used in the vanilla model above. It does not contain any attention model for the query.
\vspace{-0.05in}
\item \textbf{$\mathbf{Query}_{att}$}: This model contains the query attention model in addition to all the components in $Query_{enc}$.
\vspace{-0.05in}
\item \textbf{$\mathbf{D}_1$}: The diversity attention model as described in Section 4.1.
\vspace{-0.05in}
\item \textbf{$\mathbf{D}_2$}: The LSTM based diversity attention model as described in Section 4.1.
\vspace{-0.05in}
%%CameraReady
\item \textbf{$\mathbf{SD}_1$}: The soft diversity attention model as described in Section 4.1
\vspace{-0.05in}
\item \textbf {$\mathbf{SD}_2$}: The soft LSTM based diversity attention model as described in Section 4.1
\vspace{-0.05in}
\item \textbf {$\mathbf{B}_1$}: Diversity cell in Figure\ref{div} is replaced by the basic LSTM cell (i.e. $c_t^{diverse} = c_t$ instead of using Equation \eqref{eq12}. This helps us understand whether simply using an LSTM to track the history of context vectors (without imposing a diversity constraint) is sufficient.
\vspace{-0.05in}
\item \textbf{$\mathbf{M}_1$}: The baseline model which operates on the context vector as described in Section 5.
\item \textbf{$\mathbf{M}_2$}: The baseline model which operates on the attention weights in addition to the context vector as described in Section 5.
\end{itemize}

We used 80\% of the data for training, 10\% for validation and 10\% for testing. We create 10 such folds and report the average Rouge-1, Rouge-2, Rouge-L scores across the 10 folds. The hyperparameters (batch size and GRU cell sizes) of all the models are tuned on the validation set. We tried the following batch sizes : {32, 64} and the following GRU cell sizes {200, 300, 400}. We used Adam \cite{kingma2014adam} as the optimization algorithm with the initial learning rate set to 0.0004, $\beta_1=0.9$, $\beta_2=0.999$. We used pre-trained publicly available Glove word embeddings\footnote{http://nlp.stanford.edu/projects/glove/} and fine-tuned them during training. The same word embeddings are used for the query words and the document words.

Table \ref{res1} summarizes the results of our experiments.

\begin{center}
\begin{table}[h]
%\begin{tabular}{|p{0.1\textwidth}|p{0.1\textwidth}|p{0.1\textwidth}|p{0.1\textwidth}|}
\resizebox{\columnwidth}{!}{%
\begin{tabular}{|c|c|c|c|}
\hline
Models & ROUGE-1 & ROUGE-2 & ROUGE-L\\
\hline
Vanilla e-a-d & 13.73 &2.06 & 12.84 \\
$Query_{enc}$ & 20.87 & 3.39 & 19.38 \\
$Query_{att}$ & 29.28 & 10.24 & 28.21\\
B1 & 23.18 & 6.46 & 22.03 \\
M1 &33.06 & 13.35 &	32.17\\
M2 &18.42 & 4.47 & 17.45 \\
D1 & 33.85	& 13.65	& 32.99\\
SD1 & 31.36 & 11.23 & 30.5\\
D2 & 38.12 & 16.76 & 37.31\\
SD2 & \textbf{41.26} & \textbf{18.75} & \textbf{40.43} \\

\hline
\end{tabular}
}
\vspace{1pt}
\caption{Performance on various models using full-length ROUGE metrics}
\label{res1}
\end{table}
\end{center}

\if 0
\begin{figure}
\includegraphics[width=\columnwidth]{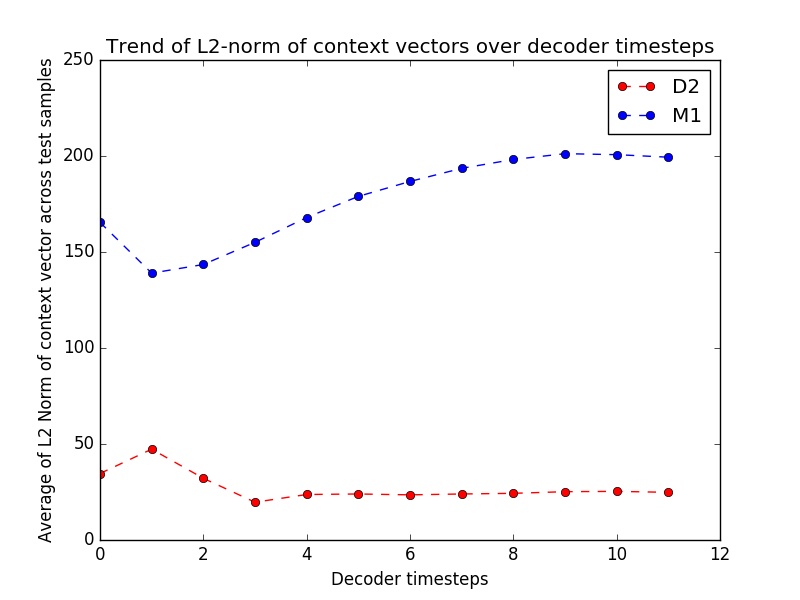}
\caption{L-2 norm of context vectors}
\label{ctx}
\end{figure}
\fi

\begin{table*}[!tbh]
\begin{center}
\begin{tabular}{|p{1\textwidth}|}
%\hline
%\textbf{Examples}\\
\hline
\textbf{Source:}Although cannabis does indeed have some harmful effects, it is no more harmful than legal substances like alcohol and tobacco. As a matter of fact, research by the British Medical Association shows that nicotine is far more addictive than cannabis. Furthermore, the consumption of alcohol and the smoking of cigarettes cause more deaths per year than does the use of cannabis (e.g. through lung cancer, stomach ulcers, accidents caused by drunk driving etc.). The legalization of cannabis will remove an anomaly in the law whereby substances that are more dangerous than cannabis are legal whilst the possession and use of cannabis remains unlawful.\\
\textbf{Query:} is marijuana harmless enough to be considered a medicine\\
\textbf{G:} marijuana is no more harmful than tobacco and alcohol\\
\boldmath${Query_{attn}}$: marijuana is no the \textbf{drug drug} for \textbf{tobacco and tobacco}\\
\textbf{D1:} marijuana is no more harmful than tobacco and tobacco\\
\textbf{SD1:} marijuana is more for evidence than tobacco and health\\
\textbf{D2:} marijuana is no more harmful than tobacco and use \\
\textbf{SD2:} marijuana is no more harmful than tobacco and alcohol\\

\hline
\textbf{Source:}Fuel cell critics point out that hydrogen is flammable, but so is gasoline. Unlike gasoline, which can pool up and burn for a long time, hydrogen dissipates rapidly. Gas tanks tend to be easily punctured, thin-walled containers, while the latest hydrogen tanks are made from Kevlar. Also, gaseous hydrogen isn't the only method of storage under consideration--BMW is looking at liquid storage while other researchers are looking at chemical compound storage, such as boron pellets.\\
\textbf{Query:} safety are hydrogen fuel cell vehicles safe\\
\textbf{G:} hydrogen in cars is less dangerous than gasoline \\
\boldmath${Query_{attn}}$:  hydrogen is \textbf{hydrogen hydrogen hydrogen} fuel energy\\
\textbf{D1:}hydrogen in cars is less natural than gasoline\\
\textbf{SD1:} hydrogen in cars is reduce risk than fuel\\
\textbf{D2:} hydrogen in waste is less effective than gasoline\\
\textbf{SD2:}hydrogen in cars is less dangerous than gasoline \\

\hline
\textbf{Source:}The basis of all animal rights should be the Golden Rule: we should treat them as we would wish them to treat us, were any other species in our dominant position.\\
\textbf{Query:}  do animals have rights that makes eating them inappropriate \\
\textbf{G:} animals should be treated as we would want to be treated   \\
\boldmath{$Query_{att}$}:
animals should be \textbf{treated} as we would protect to be \textbf{treated} \\
\textbf{D1:} animals should be \textbf{treated} as we most individual to be \textbf{treated}\\
\textbf{SD1:} animals should be \textbf{treated} as we would physically to be treated\\
\textbf{D2:} animals should be \textbf{treated} as we would illegal to be \textbf{treated}\\
\textbf{SD2:} animals should be \textbf{treated} as those would want to be \textbf{treated}\\

\hline

\end{tabular}
\vspace{1pt}
\end{center}
\caption{Summaries generated by different models. In general, we observed that the baseline models which do not use a diversity based attention model tend to produce more repetitions. Notice that the last example shows that our model is not very aggressive in dealing with the history and is able to produce valid repetitions (treated ... treated) when needed}
\label{eg}
\end{table*}

%\vspace{-0.05in}
\section{Discussions}
In this section, we discuss the results of the experiments reported in Table \ref{res1}. \\ %and \ref{eg}.\\
\noindent \textbf{1. Effect of Query:} Comparing rows 1 and 2 we observe that adding an encoder for the query and allowing it to influence the outputs of the decoder indeed improves the performance. This is expected as the query contains some keywords which could help in sharpening the focus of the summary.   

\noindent \textbf{2. Effect of Query attention model:} Comparing rows 2 and 3 we observe that using an attention model to dynamically compute the query representation at each time step improves the results. This suggests that the attention model indeed learns to focus on relevant portions of the query at different time steps.

\noindent \textbf{3. Effect of Diversity models:} All the diversity models introduced in the paper (rows 7, 8, 9, 10) give significant improvement over the non-diversity models. In particular, the modified LSTM based diversity model gives the best results. This is indeed very encouraging and Table \ref{eg} shows some sample summaries comparing the performance of different models.

\noindent \textbf{4. Comparison with baseline diversity models:} The baseline diversity model M1 performs at par with our models D1 and SD1 but not as good as D2 and SD2. However, the model M2 performs very poorly. We believe that simultaneously adding a constraint on the context vectors as well as attention weights (as is indeed the case with M2) is a bit too aggressive and leads to poor performance %note that these models were not designed for query based abstractive summarization and we simply adapted them to our task. It is not very clear why they don't perform that well but we have a few comments on this. First, the original paper itself claims that the model does not do better than a vanilla encoder-attention-decoder model when the length of the documents is small. Specifically, they show that if the length of the documents is around 100 words (6-8 sentences) then their model does not show improvements (whereas it performs well for longer documents having length $>$ 600 words). The average length of the documents in our dataset is around 66 words (4-5 sentences) which is perhaps not suitable for their model. Second, we believe that the use of LSTM and successive orthogonalization allows us to revisit topics if need be later in the summary whereas their models try to aggressively delete the history. Finally, it seems that subtracting all previous context vectors (or attention vectors) could adversely impact the magnitude of the context vectors at later time steps. To verify this, we plotted the    average L-2 norm of the context vectors obtained using $M_1$ and $D_2$. Figure \ref{ctx} shows that as the number of time steps increases the L-2 norm of the context vectors increases in the case of $M_1$ but not in the case of $D_2$. We hypothesize that this large L-2 norm perhaps causes some neurons to saturate and destabilizes the training 
(although this needs further investigation).

\noindent \textbf{5. Quantitative Analysis:} In addition to the qualitative analysis reported in Table \ref{eg} we also did a quantitative analysis by counting the number of sentences containing repeated words generated by different models. Specifically for the 1268 test instances we counted the number of sentences containing repeated words as generated by different modes. Table \ref{table_rep} summarizes this analysis. 
\begin{center}
\begin{table}[!tbh]
\begin{center}
\begin{tabular}{cc}
\hline
Model & Number\\
\hline
\textbf{$\mathbf{Query}_{attn}$} & 498\\
\textbf{$\mathbf{SD}_{1}$} & 352  \\
\textbf{$\mathbf{SD}_{2}$} & 344 \\
\textbf{$\mathbf{D}_{1}$} & 191 \\
\textbf{$\mathbf{D}_{2}$} & 179\\ 
\hline
&\\
\end{tabular}
\end{center}
\caption{Average number of sentences with repeating words across 10 folds}
\label{table_rep}
\end{table}
\end{center}

\section{Conclusion}
In this work we proposed a query-based summarization method. The unique feature of the model is
a novel diversification mechanism based on successive orthogonalization. This gives us the flexibility to: (i) provide diverse context vectors at successive time steps and (ii) pay attention to words repeatedly if need be later in the summary (as opposed to existing models which aggressively delete the history). We also introduced a new data set and empirically verified we perform significantly better (gain of 28\% (absolute) in ROUGE-L score) than applying a plain encode-attend-decode mechanism to this problem. We observe that adding an attention mechanism on the query string gives significant improvements. We also compare with a state of the art diversity  model and outperform it by a good margin (gain of 7\% (absolute) in ROUGE-L score). The diversification model proposed is general enough to apply to other NLG tasks with suitable modifications and we are currently working on extending this to dialog systems and general summarization. 

%Our model consists of the following parts (i) a basic encoder-attention-decoder model which learns to encode the document and decodes natural language sentences We first describe the basic encoder-attention-decoder model which has been used effectively for natural language generation problem   

\bibliography{acl2017}
\bibliographystyle{acl_natbib}

\end{document}